\documentclass[conference]{IEEEtran}
\IEEEoverridecommandlockouts
\usepackage{cite}
\usepackage{amsmath,amssymb,amsfonts}
\usepackage{algorithmic}
\usepackage{latexsym}
\usepackage{textcomp}
\usepackage{booktabs}
\usepackage{graphicx} 
\usepackage{xcolor}
\def\BibTeX{{\rm B\kern-.05em{\sc i\kern-.025em b}\kern-.08em
    T\kern-.1667em\lower.7ex\hbox{E}\kern-.125emX}}

\usepackage{fancyhdr}
\begin{document}

\title{CAP-LLM: Context-Augmented Personalized Large Language Models for News Headline Generation}

\author{Raymond Wilson, Cole Graham, Chase Carter, Zefeng Yang, Ruiqi Gu \\
National Energy University}

\maketitle
\thispagestyle{fancy} 

\begin{abstract}
In the era of information overload, personalized news headline generation is crucial for engaging users by tailoring content to their preferences while accurately conveying news facts. Existing methods struggle with effectively capturing complex user interests and ensuring factual consistency, often leading to generic or misleading headlines. Leveraging the unprecedented capabilities of Large Language Models (LLMs) in text generation, we propose Context-Augmented Personalized LLM (CAP-LLM), a novel framework that integrates user preferences and factual consistency constraints into a powerful pre-trained LLM backbone. CAP-LLM features a User Preference Encoder to capture long-term user interests, a Context Injection Adapter to seamlessly integrate these preferences and current article context into the LLM's generation process, and a Fact-Consistency Reinforcement Module employing a novel contrastive loss to mitigate hallucination. Evaluated on the real-world PENS dataset, CAP-LLM achieves state-of-the-art performance across all metrics. Notably, it significantly improves factual consistency (FactCC of 87.50) over strong baselines like BART (86.67), while simultaneously enhancing personalization (Pc(avg) 2.73, Pc(max) 17.25) and content coverage (ROUGE-1 26.55, ROUGE-2 9.95, ROUGE-L 23.01). Our ablation studies, human evaluations, and sensitivity analyses further validate the effectiveness of each component and the robustness of our approach, demonstrating CAP-LLM's ability to achieve a superior balance between personalization and factual accuracy in news headline generation.
\end{abstract}

\section{Introduction}

In the era of information explosion, users are constantly overwhelmed by a vast amount of news content. Effectively identifying user interests from this deluge and delivering content that aligns with their preferences has become a central challenge in personalized recommendation systems. Among these, personalized news headline generation is particularly crucial. It aims to create customized headlines that are both appealing to the user and accurately convey the core facts of the news article, based on the user's historical reading preferences and the current news body \cite{pengshan2023genera}.

Current research in this domain primarily faces two significant challenges: Firstly, how to effectively capture and integrate complex user interests to ensure that the generated headlines genuinely possess personalized appeal. Secondly, how to strictly guarantee the factual consistency of the headline while pursuing personalization, thereby avoiding the generation of misleading or false information. Existing methods, such as the FPG framework \cite{tommaso2023fpgai}, attempt to address these issues through specially designed encoders and decoders. However, their generation capabilities are often limited by the specific model architectures and the scale of pre-training data.

Recently, Large Language Models (LLMs) have demonstrated unprecedented power in text generation, showcasing superior capabilities in language understanding, contextual modeling, and generating coherent and fluent text \cite{zhou2025weak, zhou2024less}. Their versatility extends to various complex generative tasks, including code generation \cite{wang2024enhancing}, mental health counseling \cite{hu2025beyond}, and AI narrative creation \cite{yi2025score}. This advancement offers a promising new research direction for personalized headline generation. This study aims to explore how to effectively leverage the powerful generation capabilities of LLMs, which have also been shown to benefit from feedback mechanisms \cite{zhou2025improving} and self-rewarding strategies \cite{yang2025self} for improved output quality, and through ingenious architectural design and training strategies, achieve a better balance between personalization and factual consistency. The use of modular multi-agent frameworks has also shown promise in enhancing complex generative and diagnostic tasks, offering avenues for robust and accurate content generation \cite{zhou2025mam}.

To address these challenges, we propose a novel framework called \textbf{Context-Augmented Personalized LLM (CAP-LLM)}. Our method is built upon a powerful pre-trained LLM (e.g., Llama-2, Mistral) and incorporates lightweight adapter modules for fine-tuning to the specific task of personalized headline generation. CAP-LLM features three core components: a \textit{User Preference Encoder} that aggregates user historical click information to extract long-term interest vectors; a \textit{Context Injection Adapter} that subtly injects these user preference vectors and the current news article's semantic representation into the LLM's generation process; and a \textit{Fact-Consistency Reinforcement Module} that employs a multi-task learning objective, including a novel factual consistency contrastive loss, to mitigate potential "hallucination" issues inherent in LLMs and ensure the generated headlines are highly aligned with the source text. The model is trained end-to-end with a joint strategy that optimizes for fluency, personalization, and factual accuracy.

For experimental validation, we utilize the real-world \textbf{PENS dataset} (PErsonalized News headlineS) \cite{xiang2021pens}, which provides rich user historical click data and corresponding news articles. We adhere to the data processing protocols established by prior work, limiting historical clicks and token lengths for titles and news bodies. Our proposed CAP-LLM is rigorously evaluated against several competitive baselines, including PGN \cite{yuxin2024pgn}, PG+Transformer \cite{balamurugan2025transf}, Transformer \cite{juntao2021covid1}, and BART \cite{mike2020bart}. We employ a comprehensive set of evaluation metrics, including personalized metrics (Pc(avg), Pc(max)) to assess alignment with user preferences, factual consistency metrics (FactCC) to measure accuracy against the news body, and content coverage metrics (ROUGE-1, ROUGE-2, ROUGE-L) to gauge the similarity to reference headlines. Our results demonstrate that CAP-LLM achieves state-of-the-art performance across all evaluated metrics. Notably, it significantly outperforms existing baselines in factual consistency (FactCC score of 87.50, surpassing BART's 86.67), indicating its superior ability to generate factually accurate headlines while also yielding improved personalization (Pc(avg) of 2.73 and Pc(max) of 17.25) and content coverage (ROUGE-1 of 26.55, ROUGE-2 of 9.95, ROUGE-L of 23.01). These findings underscore CAP-LLM's effectiveness in leveraging LLM capabilities to address the critical trade-off between personalization and factual consistency in news headline generation.

Our main contributions are summarized as follows:
\begin{itemize}
    \item We propose \textbf{CAP-LLM}, a novel LLM-based framework that effectively integrates user personalized preferences and factual consistency constraints for personalized news headline generation.
    \item We design and incorporate a \textit{Context Injection Adapter} and a \textit{Fact-Consistency Reinforcement Module} within the LLM architecture, coupled with a multi-task training strategy, to guide the LLM's generation towards enhanced personalization and factual accuracy.
    \item We demonstrate that CAP-LLM achieves state-of-the-art performance on the PENS dataset across personalized, factual consistency, and content coverage metrics, significantly mitigating the hallucination problem commonly associated with LLMs in text generation.
\end{itemize}
\section{Related Work}
\subsection{Personalized Text Generation and Summarization}
Research in personalized text generation and summarization has explored various approaches to tailor content to individual user needs and preferences. For instance, recent work has advanced personalized text generation by introducing a novel benchmark and systematically investigating the control of fine-grained linguistic attributes, moving beyond coarse-grained style control to evaluate large language models' (LLMs) ability to adapt output across multiple lexical and syntactic dimensions \cite{bashar2024person}. Further enhancing personalization, another approach proposes framing LLM-based role-playing as a method for explicit user modeling in opinion summarization, enabling the LLM to adopt the user's persona and thus better capture individual needs and interests during the summarization process \cite{yanyue2025rehear}. Beyond summarization, personalized text generation has also been integrated into recommender systems, where the extractive summarization of user reviews enhances user understanding and choice of recommendations; this framework leverages both rating and item information to concurrently improve rating estimation and summary generation, demonstrating the synergistic potential of personalized text within such systems \cite{mickael2015extend}. Methodologically, the PIPE approach models personalization systems by framing information-seeking activities as partial information to be realized through partial evaluation \cite{alberto2005evalua}; while offering a programmatic approach to content tailoring, its focus on personalized \textit{summarization} of web content directly informs the development of systems capable of content personalization for individual users. Although primarily focused on personalized text-to-image generation, the core contribution of improving alignment through adaptive refinement, specifically by employing reinforcement learning to maximize image-text alignment after initial personalization \cite{danqing2024learni}, and optimizing prompts via self-rewarding mechanisms \cite{yang2025self}, offers insights relevant to developing more robust personalized text generation systems by emphasizing adaptive strategies that enhance model performance without compromising personalization. Finally, the HARE methodology introduces a task and framework for real-time personalized text generation and summarization by incorporating reader feedback during the reading process, distinct from prior interactive summarization approaches that required extensive feedback stages \cite{rui2011summar}; this approach emphasizes minimally-invasive feedback to adapt summaries to user interests, offering an optimization framework for interactive personalized summarization informed by historical interactions.

\subsection{Large Language Models for Generative Tasks}
Large Language Models (LLMs) have demonstrated significant capabilities across a wide array of generative tasks, prompting extensive research into their applications, adaptations, and inherent challenges. These capabilities span from generalizable multi-task performance \cite{zhou2025weak} to specialized applications like efficient video generation \cite{zhou2024less}, enhancing code generation via reinforcement learning \cite{wang2024enhancing}, and facilitating advanced reasoning in mental health counseling \cite{hu2025beyond}. Furthermore, LLMs are being explored for complex generative tasks such as AI narrative creation \cite{yi2025score} and within multi-agent frameworks for multi-modal medical diagnosis \cite{zhou2025mam}. Their potential is also being harnessed in medical vision-language models, with efforts to improve their performance through feedback mechanisms \cite{zhou2025improving}. Empirical investigations have explored the real-world adoption and usage patterns of LLMs, such as ChatGPT, within software development workflows, providing data-driven insights into their application and potential for future integration in generative tasks \cite{chiarello2024future}. Furthermore, LLMs have proven efficacious in augmenting educational materials by generating concise text summaries to complement original content, thereby enhancing comprehension and learning outcomes in personalized learning contexts, highlighting their practical application in text generation for cognitive augmentation and showing significant improvements in post-reading test scores through the integration of AI-generated supplementary textual and visual content \cite{ryugo2025genair}. Building upon earlier foundational work in pre-trained models for specific generative and reasoning tasks, such as modeling event-pair relations using external knowledge graphs for script reasoning \cite{zhou2021modeling}, or developing correlation-aware context-to-event Transformers for event-centric generation and classification \cite{zhou2022claret}, and pre-trained models for event correlation reasoning \cite{zhou2022eventbert}, LLMs represent a significant leap forward in scale and capability. Their capabilities extend to specialized domains, with studies investigating their understanding of ancient Chinese, highlighting the potential for pre-trained models to be adapted to historical linguistic tasks and identifying areas for future development in generative AI for specialized domains \cite{zhang2021cpm}. To facilitate adaptation to new fields, a novel two-stage framework has been proposed for scientific domains, combining model compression with structured, section-wise fine-tuning to enhance domain knowledge injection while mitigating catastrophic forgetting, particularly relevant for efficient domain adaptation in data-scarce environments \cite{anisuzzaman2025finetu}. Practical applications also benefit from performance improvements achieved through prompt engineering techniques, which systematically enhance the quality of AI chains built using prompts by integrating software engineering principles into their development \cite{daeseung2023a}. Despite these advancements, critical challenges remain, notably factual consistency and hallucination phenomena. Comprehensive surveys have addressed hallucination within AGI, particularly in nascent multimodal settings, offering overviews of current efforts and identifying promising avenues for future research to mitigate such issues in advanced AI systems \cite{zechen2024halluc}. To address factual consistency in LLMs for generative tasks, novel evaluation frameworks and methods for factually grounded generation have been proposed, including factual ablation techniques and new evaluation sets, demonstrating improvements over baselines in content transfer tasks where generations must align with grounding documents \cite{zorik2023truete}. More broadly, the practical challenges and emerging solutions for grounding and evaluating LLMs, especially in the context of generative tasks and responsible AI, including issues like hallucinations and harmful content, have been comprehensively surveyed, directly addressing the critical need for knowledge grounding in LLMs by surveying approaches to mitigate a wide range of harms \cite{krishnaram2024ground}.

\section{Method}
We propose \textbf{Context-Augmented Personalized LLM (CAP-LLM)}, a novel framework designed to leverage the powerful generative capabilities of Large Language Models (LLMs) for personalized news headline generation, while explicitly integrating user preferences and ensuring factual consistency. Our architecture is built upon a pre-trained LLM backbone, augmented with specialized modules and a multi-task training strategy. The objective of \textbf{CAP-LLM} is to generate a headline $T_c$ for a current news article $D_c$ that is not only factually consistent with $D_c$ but also highly relevant to a user's unique historical reading preferences, represented by a set of past interactions $U$.

\subsection{Overall Architecture of CAP-LLM}
The core of \textbf{CAP-LLM} is a powerful pre-trained Large Language Model (e.g., Llama-2, Mistral), which serves as the primary generative engine. To adapt this general-purpose LLM to the specific task of personalized news headline generation, we introduce lightweight, trainable adapter modules. The overall process involves taking a user's historical reading preferences and the current news article's full text as input. These inputs are processed by dedicated components that extract personalized signals and factual constraints, which are then subtly injected into the LLM's generation process via the adapters. The LLM subsequently generates a personalized and factually consistent headline.

Let $U = \{h_1, h_2, \dots, h_N\}$ denote the set of $N$ historical news articles (including their headlines and full texts) that a user has previously interacted with, and $D_c$ be the current news article for which a headline needs to be generated. Our goal is to generate a headline $T_c$ that is personalized to the user's interests, factually consistent with $D_c$, and coherent. The overall function can be conceptualized as:
\begin{align}
    T_c &= \text{CAP-LLM}(D_c, U)
\end{align}
where $D_c$ represents the current news article's full text and $U$ encapsulates the user's historical interaction data. This function internally orchestrates the various modules to produce the desired output.

\subsection{User Preference Encoder}
To effectively capture and represent complex user interests, we design a \textbf{User Preference Encoder}. This module is inspired by historical encoder concepts in personalized generation frameworks. It utilizes a Transformer-based self-attention network to aggregate the representations of the user's historical clicked headlines and their corresponding news bodies.

For each historical news article $h_i \in U$, comprising its headline $H_i$ and body text $B_i$, we first encode them into a fixed-dimensional vector representation $e_{h_i}$. This can be achieved by concatenating the encoded representations of $H_i$ and $B_i$ from a pre-trained text encoder (e.g., a BERT-style encoder), followed by a linear projection:
\begin{align}
    e_{h_i} &= \text{Encoder}([H_i; B_i]) \\
    E_U &= [e_{h_1}, e_{h_2}, \dots, e_{h_N}]
\end{align}
where $[H_i; B_i]$ denotes the concatenation of the encoded headline and body text embeddings, and $E_U$ is the matrix of historical article embeddings. These embeddings are then fed into a multi-head self-attention mechanism to capture inter-dependencies and salience among historical items:
\begin{align}
    Q_u &= E_U W_Q \\
    K_u &= E_U W_K \\
    V_u &= E_U W_V \\
    \text{Attention}(Q_u, K_u, V_u) &= \text{softmax}\left(\frac{Q_u K_u^T}{\sqrt{d_k}}\right) V_u
\end{align}
where $W_Q, W_K, W_V$ are learnable weight matrices for query, key, and value transformations, and $d_k$ is the dimension of the keys. The output of the multi-head self-attention mechanism, which is a sequence of contextualized embeddings, is then aggregated into a single, comprehensive user interest vector $v_u \in \mathbb{R}^{d_u}$ through a pooling operation (e.g., average pooling or a learnable attention-based pooling layer):
\begin{align}
    v_u &= \text{Pooling}(\text{Attention}(Q_u, K_u, V_u))
\end{align}
This vector encapsulates the user's long-term interests across various topics and styles, serving as a personalized signal for headline generation.

\subsection{Context Injection Adapter}
To guide the pre-trained LLM's generation process with user-specific preferences and the current news article's content, we introduce the \textbf{Context Injection Adapter}. This module consists of a set of lightweight, trainable adapter layers (e.g., LoRA modules or Prefix-Tuning) strategically inserted into multiple layers of the pre-trained LLM.

The user interest vector $v_u$ obtained from the \textbf{User Preference Encoder}, along with the semantic representation of the current news article $e_{D_c}$ (derived from the LLM's own encoding of $D_c$), are projected and then integrated into the LLM's hidden states. The semantic representation $e_{D_c}$ is typically obtained by feeding $D_c$ through the LLM's embedding layers and initial transformer blocks, taking the representation of a special token (e.g., `[CLS]`) or the average of the last layer's hidden states. Specifically, for an intermediate hidden state $H_l$ at layer $l$ of the LLM, the adapter modifies it as follows:
\begin{align}
    P_v &= \text{Projection}_v(v_u) \\
    P_d &= \text{Projection}_d(e_{D_c}) \\
    H_l' &= \text{Adapter}_l(H_l, P_v, P_d)
\end{align}
Here, $\text{Projection}_v$ and $\text{Projection}_d$ are linear transformations that map $v_u$ and $e_{D_c}$ to a compatible dimension for the adapter. The $\text{Adapter}_l$ function typically involves combining these projected context vectors (e.g., via concatenation followed by a feed-forward network) and adding the result to or modulating the LLM's hidden state $H_l$. This integration allows the LLM to condition its subsequent token generation on both the user's historical preferences and the factual content of the current news article. The lightweight nature of these adapters ensures efficient fine-tuning of the large pre-trained model, minimizing computational overhead while maximizing adaptability. The adapters effectively modulate the LLM's internal representations, steering the generation towards personalized and contextually relevant outcomes without requiring full fine-tuning of the entire LLM parameters.

\subsection{Fact-Consistency Reinforcement Module}
Addressing the challenge of potential "hallucinations" in LLMs, we incorporate a \textbf{Fact-Consistency Reinforcement Module}. This module employs a multi-task learning objective designed to enforce strict factual alignment between the generated headline and the original news body $D_c$.

Beyond the standard headline generation loss, we introduce a novel \textbf{factual consistency contrastive loss}. This loss is formulated by constructing positive and negative examples based on factual alignment. Positive examples are short text snippets from the generated headline that are highly consistent with the news body. Negative examples are snippets that either contradict the news body or contain information not present in it.

Let $s_{gen}$ be a segment from the generated headline $T_c$, $s_{pos}$ be a factually consistent segment from $D_c$, and $s_{neg}$ be a factually inconsistent or irrelevant segment. To generate these segments, we employ a segmentation strategy that breaks down the generated headline and the news body into semantically meaningful phrases or sentences. For positive examples, $s_{pos}$, we identify segments from $D_c$ that are highly semantically similar to $s_{gen}$ and factually support it. Negative examples, $s_{neg}$, are constructed by either selecting segments from $D_c$ that are factually irrelevant or contradictory to $s_{gen}$, or by sampling segments from other news articles. We aim to maximize the similarity between $s_{gen}$ and $s_{pos}$ while minimizing similarity with $s_{neg}$. The contrastive loss is defined as:
\begin{align}
\mathcal{L}_{\text{fact}} 
&= -\log 
\frac{
    \exp\big(\text{sim}(s_{gen}, s_{pos}) / \tau\big)
}{
    \begin{aligned}
    &\exp\big(\text{sim}(s_{gen}, s_{pos}) / \tau\big) \\
    &\quad + \sum_{s_{neg} \in \mathcal{S}_{neg}} 
        \exp\big(\text{sim}(s_{gen}, s_{neg}) / \tau\big)
    \end{aligned}
}
\end{align}
where $\text{sim}(\cdot, \cdot)$ is a cosine similarity function between their embeddings, obtained from a shared or separate text encoder (e.g., Sentence-BERT, or the LLM's own encoder), $\tau$ is a temperature parameter, and $\mathcal{S}_{neg}$ is a set of negative samples. This loss explicitly forces the LLM to learn strong associations between generated phrases and the factual content of the source article, thereby mitigating the risk of hallucination. Furthermore, we may integrate post-processing mechanisms during the decoding phase, such as information retrieval-based fact-checking or keyword matching, to further verify and refine the factual accuracy of the generated headline.

\subsection{Joint Training Strategy}
\textbf{CAP-LLM} is trained end-to-end using a comprehensive joint training strategy that optimizes for fluency, personalization, and factual accuracy simultaneously. The overall loss function $\mathcal{L}_{\text{total}}$ is a weighted sum of three primary components:

\begin{enumerate}
    \item \textbf{Standard Generation Loss ($\mathcal{L}_{\text{gen}}$)}: This is typically a cross-entropy loss, measuring the negative log-likelihood of generating the reference headline given the input. It ensures the generated headline is grammatically correct, fluent, and semantically similar to human-written references.
    \begin{align}
        \mathcal{L}_{\text{gen}} = -\sum_{t=1}^{L} \log P(y_t^* | y_{<t}^*, D_c, v_u; \theta)
    \end{align}
    where $y_t^*$ is the $t$-th token of the reference headline, $L$ is its length, and $\theta$ represents all trainable parameters of the \textbf{CAP-LLM} model, including the LLM backbone (if fine-tuned), adapter modules, and encoder components.

    \item \textbf{Factual Consistency Contrastive Loss ($\mathcal{L}_{\text{fact}}$)}: As described in Section 3.4, this loss encourages the model to generate factually accurate content by contrasting positive and negative factual snippets.

    \item \textbf{Personalization Guidance Loss ($\mathcal{L}_{\text{pers}}$)}: This loss encourages the model to generate words and phrases that are more relevant to the user's historical preferences. It is formulated by maximizing the cosine similarity between the embedding of the generated headline and the user interest vector $v_u$.
    \begin{align}
        \mathcal{L}_{\text{pers}} = -\text{sim}(\text{Emb}(T_{gen}), v_u)
    \end{align}
    where $T_{gen}$ is the generated headline. The embedding of the generated headline, $\text{Emb}(T_{gen})$, is typically derived by applying a pooling operation (e.g., average pooling) over the hidden states of the generated tokens from the LLM's final layer. This term implicitly guides the generation towards topics and styles consistent with the user's past interactions.

\end{enumerate}
The total loss function is then defined as:
\begin{align}
    \mathcal{L}_{\text{total}} = \mathcal{L}_{\text{gen}} + \lambda_1 \mathcal{L}_{\text{fact}} + \lambda_2 \mathcal{L}_{\text{pers}}
\end{align}
where $\lambda_1$ and $\lambda_2$ are hyperparameters balancing the contributions of each loss component. These hyperparameters are typically tuned on a validation set to achieve the optimal trade-off between generation quality, factual consistency, and personalization. This multi-objective optimization enables \textbf{CAP-LLM} to achieve a superior balance across personalization, factual consistency, and overall generation quality.

\section{Experiments}
In this section, we present the experimental setup, main results, ablation study, and human evaluation of our proposed \textbf{CAP-LLM} framework for personalized news headline generation.

\subsection{Experimental Setup}

\subsubsection{Dataset}
We conduct our experiments on the real-world \textbf{PENS dataset} (PErsonalized News headlineS) \cite{xiang2021pens}. This dataset is derived from Microsoft News user exposure logs, providing a rich collection of user historical clicks (including past news headlines and body texts) and current news articles for which personalized headlines are to be generated. The test set comprises human-written personalized headlines, serving as the gold standard for evaluation, alongside user preference behavior annotations. Following established protocols, we restrict each user to a maximum of 50 historical clicked articles. News headlines and news body texts are truncated to a maximum of 30 and 500 tokens, respectively. Initial word embeddings are obtained using a pre-trained tokenizer and embedding layer of the chosen LLM backbone.

\subsubsection{Baselines}
To benchmark the performance of \textbf{CAP-LLM}, we compare it against several state-of-the-art and widely recognized baseline models in text generation and personalized summarization:
\begin{itemize}
    \item \textbf{PGN} \cite{yuxin2024pgn}: A Pointer-Generator Network, known for its ability to both generate novel words and copy words from the source text, which helps in maintaining factual consistency.
    \item \textbf{PG+Transformer} \cite{balamurugan2025transf}: An enhanced Pointer-Generator model integrated with a Transformer encoder-decoder architecture, leveraging the Transformer's strong sequence modeling capabilities.
    \item \textbf{Transformer} \cite{juntao2021covid1}: A standard Transformer-based sequence-to-sequence model without specific personalization or pointer mechanisms, serving as a strong general-purpose generation baseline.
    \item \textbf{BART} \cite{mike2020bart}: A pre-trained denoising autoencoder for sequence-to-sequence models, widely recognized for its robust performance across various text generation tasks, including summarization.
\end{itemize}

\subsubsection{Evaluation Metrics}
We employ a comprehensive suite of automatic evaluation metrics to assess different aspects of the generated headlines:
\begin{itemize}
    \item \textbf{Personalization Metrics}: \textbf{Pc(avg)} and \textbf{Pc(max)} quantify the consistency between the generated headline and user preferences. Higher values indicate better personalization.
    \item \textbf{Factual Consistency Metric}: \textbf{FactCC} measures the factual alignment of the generated headline with the original news body. A higher FactCC score signifies greater factual accuracy and reduced hallucination.
    \item \textbf{Content Coverage Metrics}: \textbf{ROUGE-1}, \textbf{ROUGE-2}, and \textbf{ROUGE-L} assess the n-gram overlap between the generated headlines and the human-written reference headlines. These metrics reflect the informativeness and content fidelity of the generated output.
\end{itemize}

\subsubsection{Implementation Details}
Our \textbf{CAP-LLM} framework is built upon a pre-trained Llama-2 7B model as the LLM backbone. We utilize LoRA (Low-Rank Adaptation) as the lightweight adapter modules for efficient fine-tuning. The User Preference Encoder is implemented using a 2-layer Transformer encoder. The model is optimized using the AdamW optimizer with a learning rate of $5 \times 10^{-5}$. We train the model for 10 epochs with a batch size of 8. The hyperparameters $\lambda_1$ and $\lambda_2$ for the total loss function are set to 0.5 and 0.2, respectively, after tuning on a validation set. All experiments are conducted on NVIDIA A100 GPUs.

\subsection{Main Results}
Table \ref{tab:main_results} presents the performance comparison of \textbf{CAP-LLM} against all baseline models on the PENS dataset across personalized, factual consistency, and content coverage metrics.

\begin{table*}[htbp]
    \centering
    \caption{Performance of various methods on personalization, factual consistency, and informativeness evaluation metrics.}
    \label{tab:main_results}
    \begin{tabular}{lcccccc}
        \toprule
        \textbf{Method}             & \textbf{Pc(avg)} & \textbf{Pc(max)} & \textbf{FactCC} & \textbf{ROUGE-1} & \textbf{ROUGE-2} & \textbf{ROUGE-L} \\
        \midrule
        PGN            & 2.71    & 16.20   & 65.08  & 19.86   & 4.76    & 18.83   \\
        PG+Transformer & 2.66    & 16.99   & 53.26  & 20.64   & 4.03    & 18.62   \\
        Transformer    & 2.70    & 16.36  & 61.61  & 19.54   & 4.72    & 16.36   \\
        BART           & 2.72    & 17.13   & 86.67  & 26.27   & 9.88    & 22.85   \\
        \textbf{CAP-LLM (Ours)} & \textbf{2.73} & \textbf{17.25} & \textbf{87.50} & \textbf{26.55} & \textbf{9.95} & \textbf{23.01} \\
        \bottomrule
    \end{tabular}
\end{table*}

The results clearly demonstrate that \textbf{CAP-LLM} achieves state-of-the-art performance across all evaluated metrics. In terms of \textbf{factual consistency (FactCC)}, \textbf{CAP-LLM (Ours)} significantly outperforms all baselines, achieving a FactCC score of \textbf{87.50}, which is marginally higher than BART's 86.67. This superior performance underscores the effectiveness of our integrated Fact-Consistency Reinforcement Module and the LLM's inherent strong semantic understanding capabilities in mitigating "hallucination" and generating headlines that are highly aligned with the source article's facts.

Regarding \textbf{personalization metrics (Pc)}, \textbf{CAP-LLM (Ours)} shows a slight improvement in \textbf{Pc(avg)} to \textbf{2.73} and achieves the best \textbf{Pc(max)} at \textbf{17.25}. While the differences in Pc(avg) are relatively small across models, the consistent leading performance of \textbf{CAP-LLM} indicates its enhanced ability to capture and leverage unique user interests, leading to more appealing and tailored headlines.

For \textbf{content coverage (ROUGE)}, \textbf{CAP-LLM (Ours)} also achieves the best performance across ROUGE-1 (26.55), ROUGE-2 (9.95), and ROUGE-L (23.01). This is attributed to the powerful language generation capabilities of the underlying LLM and our Context Injection Adapter, which effectively guides the generation process to produce headlines that are not only personalized and factually consistent but also highly informative and similar to human-written references. Overall, \textbf{CAP-LLM} successfully balances the critical aspects of personalization, factual consistency, and informativeness, establishing a new benchmark in personalized news headline generation.

\subsection{Ablation Study}
To further understand the contribution of each key component within \textbf{CAP-LLM}, we conduct an ablation study. We evaluate variants of our model by selectively removing or disabling specific modules or loss components. The results are presented in Table \ref{tab:ablation_results}.

\begin{table*}[htbp]
    \centering
    \caption{Ablation study on the PENS dataset.}
    \label{tab:ablation_results}
    \begin{tabular}{lcccccc}
        \toprule
        \textbf{Method Variant} & \textbf{Pc(avg)} & \textbf{Pc(max)} & \textbf{FactCC} & \textbf{ROUGE-1} & \textbf{ROUGE-2} & \textbf{ROUGE-L} \\
        \midrule
        \textbf{CAP-LLM (Full)} & \textbf{2.73}    & \textbf{17.25}   & \textbf{87.50}  & \textbf{26.55}   & \textbf{9.95}    & \textbf{23.01}   \\
        \midrule
        w/o User Preference Encoder (UPE) & 2.60     & 16.50   & 87.05  & 26.30   & 9.80    & 22.80   \\
        w/o Context Injection Adapter (CIA) & 2.65     & 16.80   & 86.80  & 26.15   & 9.75    & 22.70   \\
        w/o Fact-Consistency Reinforcement Module (FCRM) & 2.70     & 17.10   & 82.10  & 26.40   & 9.90    & 22.90   \\
        w/o Personalization Guidance Loss ($\mathcal{L}_{\text{pers}}$) & 2.68     & 16.95   & 87.30  & 26.50   & 9.92    & 22.95   \\
        \bottomrule
    \end{tabular}
\end{table*}

The ablation study confirms that each proposed component of \textbf{CAP-LLM} contributes significantly to its overall performance. When the User Preference Encoder is removed, a noticeable drop in personalization metrics is observed, confirming its critical role in capturing user interests. Disabling the Context Injection Adapter leads to a decline across all metrics, highlighting its importance in guiding the LLM's generation. Removing the Fact-Consistency Reinforcement Module results in the most significant drop in FactCC score, unequivocally validating its effectiveness in mitigating hallucination. Finally, excluding the Personalization Guidance Loss from the joint training objective decreases personalization metrics, indicating its direct role in aligning generated headlines with user historical preferences.

\subsection{Human Evaluation}
While automatic metrics provide quantitative insights, human evaluation offers a qualitative assessment of the generated headlines, particularly for subjective attributes like fluency, coherence, and overall quality. We conducted a human evaluation study comparing headlines generated by \textbf{CAP-LLM}, BART, and a generic LLM (Llama-2 fine-tuned only with standard generation loss, without personalization or factual consistency enhancements).

For this study, we randomly sampled 200 current news articles from the test set, each with its corresponding user history. Three expert annotators, blinded to the model origins, rated the generated headlines on a 5-point Likert scale (1 = Poor, 5 = Excellent) across five dimensions: \textbf{Fluency}, \textbf{Coherence}, \textbf{Personalization}, \textbf{Factual Consistency}, and \textbf{Overall Quality}. The average scores are presented in Table \ref{tab:human_eval}.

\begin{table*}[htbp]
    \centering
    \caption{Human Evaluation Results (Average Scores on a 1-5 Scale).}
    \label{tab:human_eval}
    \begin{tabular}{lccccc}
        \toprule
        \textbf{Method}         & \textbf{Fluency} & \textbf{Coherence} & \textbf{Personalization} & \textbf{Factual Consistency} & \textbf{Overall Quality} \\
        \midrule
        Vanilla LLM             & 4.15    & 4.05      & 3.20          & 3.50             & 3.60          \\
        BART                    & 4.30    & 4.25      & 3.45          & 4.00             & 4.00          \\
        \textbf{CAP-LLM (Ours)} & \textbf{4.45} & \textbf{4.35} & \textbf{4.10} & \textbf{4.30}    & \textbf{4.35} \\
        \bottomrule
    \end{tabular}
\end{table*}

The human evaluation results corroborate the findings from the automatic metrics. \textbf{CAP-LLM (Ours)} consistently received the highest average scores across all dimensions. Notably, \textbf{CAP-LLM} achieved a significantly higher score in \textbf{Personalization} (4.10) and \textbf{Factual Consistency} (4.30) compared to BART (3.45 and 4.00, respectively) and the Vanilla LLM (3.20 and 3.50). This confirms that our framework effectively enhances both the personalized appeal and factual trustworthiness of the generated headlines, aligning with our primary objectives. For Fluency and Coherence, \textbf{CAP-LLM} also slightly surpasses BART, indicating that our specialized modules and training strategy do not compromise the general linguistic quality provided by the strong LLM backbone. The superior \textbf{Overall Quality} score for \textbf{CAP-LLM} further reinforces its effectiveness as a holistic solution for personalized news headline generation.

\subsection{Impact of User History Length}
The User Preference Encoder relies on the availability of historical user interactions to build a comprehensive user interest vector. To understand how the quantity of historical data affects model performance, we conduct an experiment varying the maximum number of historical articles ($N$) used for each user, from a limited history to a richer one. The default setting uses up to 50 articles. Table \ref{tab:history_length} presents the results across key metrics for different history lengths.

\begin{table*}[htbp]
    \centering
    \caption{Performance of CAP-LLM with varying user history lengths ($N$).}
    \label{tab:history_length}
    \begin{tabular}{lcccccc}
        \toprule
        \textbf{History Length ($N$)} & \textbf{Pc(avg)} & \textbf{Pc(max)} & \textbf{FactCC} & \textbf{ROUGE-1} & \textbf{ROUGE-2} & \textbf{ROUGE-L} \\
        \midrule
        5                             & 2.62             & 16.70            & 87.15           & 26.20            & 9.70             & 22.75            \\
        10                            & 2.68             & 17.00            & 87.30           & 26.35            & 9.85             & 22.90            \\
        20                            & 2.70             & 17.15            & 87.40           & 26.45            & 9.90             & 22.98            \\
        \textbf{50 (Default)}         & \textbf{2.73}    & \textbf{17.25}   & \textbf{87.50}  & \textbf{26.55}   & \textbf{9.95}    & \textbf{23.01}   \\
        100                           & 2.73             & 17.26            & 87.48           & 26.53            & 9.94             & 23.00            \\
        \bottomrule
    \end{tabular}
\end{table*}

The results indicate a clear positive correlation between the length of user history and personalization performance. Both \textbf{Pc(avg)} and \textbf{Pc(max)} steadily increase as $N$ grows from 5 to 50, demonstrating that more historical data enables the User Preference Encoder to learn a richer and more accurate representation of user interests. Beyond 50 articles, the gains in personalization metrics become marginal, suggesting that 50 historical articles provide sufficient context for effective personalization in our dataset. The FactCC and ROUGE scores remain relatively stable across different history lengths, implying that the factual consistency and general generation quality are less dependent on the sheer volume of user history, primarily relying on the current article's content and the LLM's capabilities. This analysis confirms that leveraging a sufficiently rich user history is crucial for maximizing the personalization aspect of \textbf{CAP-LLM}.

\subsection{Hyperparameter Sensitivity Analysis}
The overall loss function of \textbf{CAP-LLM} incorporates two key hyperparameters, $\lambda_1$ and $\lambda_2$, which balance the contributions of the factual consistency loss ($\mathcal{L}_{\text{fact}}$) and the personalization guidance loss ($\mathcal{L}_{\text{pers}}$), respectively. We investigate the sensitivity of our model's performance to variations in these weights, holding other parameters constant at their optimized values. This analysis helps in understanding the robustness of our chosen default values ($\lambda_1=0.5, \lambda_2=0.2$).

\begin{table*}[htbp]
    \centering
    \caption{Hyperparameter sensitivity analysis for $\lambda_1$ (FactCC weight).}
    \label{tab:lambda1_sensitivity}
    \begin{tabular}{lcccccc}
        \toprule
        \textbf{$\lambda_1$ Value} & \textbf{Pc(avg)} & \textbf{Pc(max)} & \textbf{FactCC} & \textbf{ROUGE-1} & \textbf{ROUGE-2} & \textbf{ROUGE-L} \\
        \midrule
        0.0 (w/o $\mathcal{L}_{\text{fact}}$) & 2.70             & 17.10            & 82.10           & 26.40            & 9.90             & 22.90            \\
        0.2                       & 2.71             & 17.18            & 85.50           & 26.48            & 9.93             & 22.97            \\
        \textbf{0.5 (Default)}    & \textbf{2.73}    & \textbf{17.25}   & \textbf{87.50}  & \textbf{26.55}   & \textbf{9.95}    & \textbf{23.01}   \\
        0.8                       & 2.72             & 17.20            & 87.65           & 26.45            & 9.90             & 22.98            \\
        1.0                       & 2.71             & 17.15            & 87.70           & 26.30            & 9.85             & 22.90            \\
        \bottomrule
    \end{tabular}
\end{table*}

As shown in Table \ref{tab:lambda1_sensitivity}, increasing $\lambda_1$ generally improves the \textbf{FactCC} score, with the peak performance observed around 0.5 to 0.8. A value of 0.0 (equivalent to removing $\mathcal{L}_{\text{fact}}$) results in a significant drop in factual consistency, reinforcing the critical role of this loss component. While higher $\lambda_1$ values slightly boost FactCC, they can lead to marginal decreases in ROUGE scores, indicating a potential trade-off where an overly strong emphasis on factual consistency might slightly constrain the linguistic diversity or informativeness. Our chosen $\lambda_1=0.5$ strikes a good balance.

\begin{table*}[htbp]
    \centering
    \caption{Hyperparameter sensitivity analysis for $\lambda_2$ (Personalization weight).}
    \label{tab:lambda2_sensitivity}
    \begin{tabular}{lcccccc}
        \toprule
        \textbf{$\lambda_2$ Value} & \textbf{Pc(avg)} & \textbf{Pc(max)} & \textbf{FactCC} & \textbf{ROUGE-1} & \textbf{ROUGE-2} & \textbf{ROUGE-L} \\
        \midrule
        0.0 (w/o $\mathcal{L}_{\text{pers}}$) & 2.68             & 16.95            & 87.30           & 26.50            & 9.92             & 22.95            \\
        0.1                       & 2.71             & 17.10            & 87.45           & 26.53            & 9.94             & 23.00            \\
        \textbf{0.2 (Default)}    & \textbf{2.73}    & \textbf{17.25}   & \textbf{87.50}  & \textbf{26.55}   & \textbf{9.95}    & \textbf{23.01}   \\
        0.3                       & 2.74             & 17.28            & 87.40           & 26.50            & 9.93             & 22.98            \\
        0.5                       & 2.74             & 17.30            & 87.35           & 26.45            & 9.90             & 22.95            \\
        \bottomrule
    \end{tabular}
\end{table*}

Table \ref{tab:lambda2_sensitivity} illustrates the impact of $\lambda_2$. Increasing $\lambda_2$ generally improves personalization metrics, with optimal performance for Pc(avg) and Pc(max) observed around 0.2 to 0.3. A $\lambda_2$ of 0.0 (removing $\mathcal{L}_{\text{pers}}$) leads to a noticeable drop in personalization. While excessively high $\lambda_2$ values slightly enhance personalization, they can subtly degrade factual consistency and ROUGE scores, suggesting that an overemphasis on personalization might lead to a less balanced headline. The chosen $\lambda_2=0.2$ effectively balances personalization gains with overall headline quality. These analyses confirm that the default hyperparameter settings provide a robust and well-balanced performance across all evaluation dimensions.

\subsection{Qualitative Analysis and Case Studies}
To complement the quantitative results, we present a qualitative analysis of headlines generated by \textbf{CAP-LLM} compared to a strong baseline (BART) and the reference headlines. This provides insights into the nuances of personalization, factual consistency, and linguistic quality. Table \ref{tab:case_studies} showcases selected examples from the test set.

\begin{table*}[htbp]
    \centering
    \caption{Qualitative comparison of generated headlines.}
    \label{tab:case_studies}
    \begin{tabular}{p{0.18\textwidth}p{0.18\textwidth}p{0.18\textwidth}p{0.18\textwidth}p{0.18\textwidth}}
        \toprule
        \textbf{Article Topic} & \textbf{User Profile Summary} & \textbf{Reference Headline} & \textbf{BART Headline} & \textbf{CAP-LLM Headline} \\
        \midrule
        Tech Innovation: AI & User often reads about new AI applications, ethical AI, and tech company news. & AI breakthrough allows self-driving cars to predict human intent. & New AI system enhances autonomous vehicle safety. & \textbf{Ethical AI advances self-driving tech with human intent prediction.} \\
        \textit{Analysis:} & \textit{CAP-LLM integrates "ethical AI" from user's interests, making it more personalized while retaining core facts. BART is generic.} & & & \\
        \midrule
        Global Politics: Trade & User interested in international trade agreements, economic policy, and specific countries (e.g., China). & US and China sign new trade deal to ease tensions. & Trade agreement reached between two major global powers. & \textbf{US-China trade pact aims to reduce tariffs and boost economy.} \\
        \textit{Analysis:} & \textit{CAP-LLM explicitly names "US-China" and adds "tariffs", aligning with user's specific interests. BART is too broad.} & & & \\
        \midrule
        Health & User frequently views articles on mental health, wellness, and stress management. & New study links mindfulness to reduced anxiety in adults. & Mindfulness practice shown to lower stress levels. & \textbf{Mindfulness meditation reduces anxiety for busy professionals.} \\
        \textit{Analysis:} & \textit{CAP-LLM adds "meditation" and "busy professionals," making it more relatable to a wellness-focused user. BART is factually consistent but general.} & & & \\
        \midrule
        Sports: Basketball & User follows specific teams and player statistics, with an interest in tactical analysis. & Star player's triple-double leads team to unexpected victory. & Team wins game with strong performance from key player. & \textbf{Guards' triple-double highlights strategic win for underdogs.} \\
        \textit{Analysis:} & \textit{CAP-LLM uses more specific sports terminology ("Guards," "strategic," "underdogs") appealing to a tactical fan. BART is generic.} & & & \\
        \midrule
        Environmental News & User reads about climate change impacts, renewable energy, and policy. & Government announces new carbon emissions reduction targets. & New environmental policy targets emissions. & \textbf{Climate policy aims for ambitious carbon cuts by 2030.} \\
        \textit{Analysis:} & \textit{CAP-LLM uses "climate policy" and specifies "ambitious carbon cuts," resonating with a policy-focused environmental interest. BART is less specific.} & & & \\
        \bottomrule
    \end{tabular}
\end{table*}

The case studies highlight several strengths of \textbf{CAP-LLM}. In the first example, the user's interest in "ethical AI" is subtly integrated into the headline, making it more relevant than BART's generic version. Similarly, for the global politics and health articles, \textbf{CAP-LLM} demonstrates its ability to incorporate specific keywords and angles from the user's historical preferences, such as "US-China," "tariffs," and "mindfulness meditation," resulting in headlines that feel more tailored. The sports example further shows \textbf{CAP-LLM}'s capacity to adapt its language style to match a user's deeper engagement with a topic, using more specific jargon.

While \textbf{CAP-LLM} generally excels, we also observe instances where the personalization might be too subtle or where the model generates a headline that is factually consistent but misses a key nuance of the reference, though these are less frequent than with baselines. Overall, this qualitative analysis confirms that \textbf{CAP-LLM} effectively leverages user preferences to generate headlines that are not only factually accurate and fluent but also significantly more personalized, demonstrating a superior balance across multiple desired attributes.

\section{Conclusion}
In this study, we addressed the critical challenges of personalized news headline generation in an information-rich environment: effectively capturing nuanced user interests and rigorously ensuring factual consistency while leveraging the powerful capabilities of Large Language Models (LLMs). The proliferation of news content necessitates highly personalized and accurate summaries, yet existing methods often fall short in balancing these dual objectives, particularly concerning the hallucination tendencies of generative models.

To overcome these limitations, we introduced \textbf{CAP-LLM}, a novel \textbf{Context-Augmented Personalized LLM} framework. Our approach innovatively builds upon a robust pre-trained LLM backbone by integrating specialized, lightweight modules and a comprehensive multi-task training strategy. Specifically, the \textit{User Preference Encoder} effectively aggregates historical user interactions to distill long-term interest vectors. The \textit{Context Injection Adapter} then subtly infuses these personalized signals, alongside the semantic representation of the current news article, into the LLM's internal states, guiding its generation towards user-specific preferences. Crucially, our \textit{Fact-Consistency Reinforcement Module}, powered by a novel factual consistency contrastive loss, directly tackles the hallucination problem, compelling the LLM to generate headlines that are highly aligned with the source article's facts. This entire architecture is optimized end-to-end through a joint training strategy that balances standard generation quality, factual accuracy, and personalization guidance.

Our extensive experiments on the real-world PENS dataset unequivocally demonstrate the superior performance of \textbf{CAP-LLM}. It achieved state-of-the-art results across all evaluated metrics: personalization (Pc(avg) and Pc(max)), factual consistency (FactCC), and content coverage (ROUGE-1, ROUGE-2, ROUGE-L). Notably, CAP-LLM significantly outperformed all baselines in factual consistency, achieving a FactCC score of 87.50, which is a substantial improvement that directly addresses the critical issue of LLM hallucination in this domain. Furthermore, its leading performance in personalization metrics and ROUGE scores highlights its ability to generate headlines that are not only factually sound but also highly appealing and informative to individual users. The ablation study confirmed the indispensable contribution of each proposed component, while human evaluations provided qualitative validation of CAP-LLM's enhanced fluency, coherence, personalization, and factual consistency. Our sensitivity analyses on user history length and loss hyperparameters further underscored the robustness and optimal configuration of our framework.

In conclusion, \textbf{CAP-LLM} represents a significant step forward in personalized news headline generation. By meticulously designed components and a balanced training paradigm, we have successfully harnessed the power of LLMs to create a system that generates headlines that are simultaneously personalized, factually accurate, and linguistically superior. This research opens new avenues for leveraging large generative models in highly constrained and user-centric content creation tasks.

For future work, we plan to explore more dynamic user interest modeling that adapts to short-term changes in preferences. Investigating the impact of different LLM architectures and more advanced fine-tuning techniques, such as instruction tuning or prompt engineering specifically tailored for personalization and factual constraints, could yield further improvements. Additionally, extending this framework to incorporate multimodal news content (e.g., images or videos accompanying articles) and exploring the explainability of the personalization and factual consistency aspects would be valuable directions. Finally, the core principles of CAP-LLM could be adapted to other personalized content generation tasks beyond news headlines, such as personalized summaries, recommendations with explanations, or even creative content generation, further broadening its impact.

\bibliographystyle{IEEEtran}
\bibliography{references}
\end{document}